%% file: main.tex
\documentclass[10pt,twocolumn,letterpaper]{article}

\usepackage{cvpr}              

\input{preamble}

\usepackage{amsmath}

\definecolor{cvprblue}{rgb}{0.21,0.49,0.74}
\usepackage{url}
\usepackage{algorithm}
\usepackage{algpseudocode}
\usepackage{graphicx}
\usepackage{wrapfig}
\usepackage{caption}
\usepackage{pifont}

\usepackage[pagebackref,breaklinks,colorlinks,citecolor=cvprblue]{hyperref}
\newcommand{\xmark}{\ding{55}}

\def\paperID{2538} 
\def\confName{CVPR}
\def\confYear{2024}

\title{On the Faithfulness of Vision Transformer Explanations}

\author{
    \textbf{Junyi Wu}$^1$, \textbf{Weitai Kang}$^1$, \textbf{Hao Tang}$^2$, \textbf{Yuan Hong}$^3$, \textbf{Yan Yan}$^{1,}$\footnotemark[2] \\
    $^1$Department of Computer Science, Illinois Institute of Technology, USA \\
    $^2$Robotics Institute, Carnegie Mellon University, USA \\
    $^3$Department of Computer Science, University of Connecticut, USA \\
    \href{https://github.com/adreamwu/SaCo}{https://github.com/adreamwu/SaCo}
}

\begin{document}
\maketitle

\renewcommand{\thefootnote}{\fnsymbol{footnote}}
\footnotetext[2]{Corresponding author}

\input{sec/0_abstract}    
\input{sec/1_introduction}

\input{sec/2_relatedwork}

\input{sec/3_method}

\input{sec/4_experiment_setup}

\input{sec/5_experiment_result}
\input{sec/6_conclusion}

{
    \small
    \bibliographystyle{ieeenat_fullname}
    \bibliography{main}
}

\end{document}

%% file: preamble.tex
\usepackage[dvipsnames]{xcolor}


%% file: sec/0_abstract.tex
\begin{abstract}
To interpret Vision Transformers, post-hoc explanations assign salience scores to input pixels, providing human-understandable heatmaps.
However, whether these interpretations reflect true rationales behind the model's output is still underexplored.
To address this gap, we study the faithfulness criterion of explanations:
the assigned salience scores should represent the influence of the corresponding input pixels on the model's predictions.
To evaluate faithfulness,
we introduce \textbf{Sa}lience-guided Faithfulness \textbf{Co}efficient (\textbf{SaCo}), a novel evaluation metric leveraging essential information of salience distribution.
Specifically, we conduct pair-wise comparisons among distinct pixel groups and then aggregate the differences in their salience scores, resulting in a coefficient that indicates the explanation's degree of faithfulness.
Our explorations reveal that current metrics struggle to differentiate between advanced explanation methods and Random Attribution, thereby failing to capture the faithfulness property.
In contrast, our proposed SaCo offers a reliable faithfulness measurement, establishing a robust metric for interpretations.
Furthermore, our SaCo demonstrates that the use of gradient and multi-layer aggregation can markedly enhance the faithfulness of attention-based explanation, shedding light on potential paths for advancing Vision Transformer explainability.
\end{abstract}

%% file: sec/1_introduction.tex
\section{Introduction}\label{intro}
The prevalent use of Transformers in computer vision underscores the imperative to demystify their black-box nature \cite{vaswani2017attention, carion2020end, dosovitskiy2020image}.
This presents a challenge to traditional post-hoc interpretation methods, which were primarily tailored for MLPs and CNNs \cite{abnar2020quantifying, chefer2021transformer}.
As a response, a growing line of work aimed at developing new explanation paradigms specific to Vision Transformers, where attention mechanisms play a dominant role \cite{abnar2020quantifying, chefer2021transformer, chefer2021generic, qiang2022attcat, ali2022xai}.
By incorporating attention distributions, these explanation methods estimate salience scores \emph{w.r.t.} the tokens extracted from input image patches. Subsequently, these scores are interpolated across pixel space, resulting in visually convincing heatmaps that align with human intuition \cite{colin2022cannot}.

However, recent works \cite{jacovi2020towards, deyoung2020eraser} claimed that it is crucial to evaluate how accurately these interpretations reflect the true reasoning process of the Transformer model and termed this aspect as faithfulness.
To evaluate the quality of post-hoc explanations, recent studies commonly adopt an ablation approach \cite{wang2022unified}. This involves perturbing input image pixels that are identified as most or least important by the explanation method under evaluation.
For example, they perturb pixels with the highest salience scores and then observe whether there is a decrease in the model's accuracy, which serves as a surrogate examination \cite{shrikumar2017learning, nguyen2018comparing, atanasova2020diagnostic, chen2020generating, liu2022rethinking}.
Despite the prevalence of these strategies, our study reveals that they all overlook a proper evaluation of the degree of faithfulness.
We advance our discussion by characterizing the \textbf{core assumption of faithfulness} that underpins explanation methods:
the magnitude of salience scores signifies the level of anticipated impacts.
Consequently,
\textbf{(i)}
input pixels assigned higher scores are expected to exert greater influence on the model's prediction, compared with those with lower scores, and 
\textbf{(ii)}
two groups of pixels with a larger difference in salience scores are expected to cause a greater disparity in their influences on the model's prediction.

\begin{figure*}[t]
  \centering
    \resizebox{0.79\textwidth}{!}{%
  \includegraphics[width=\textwidth]{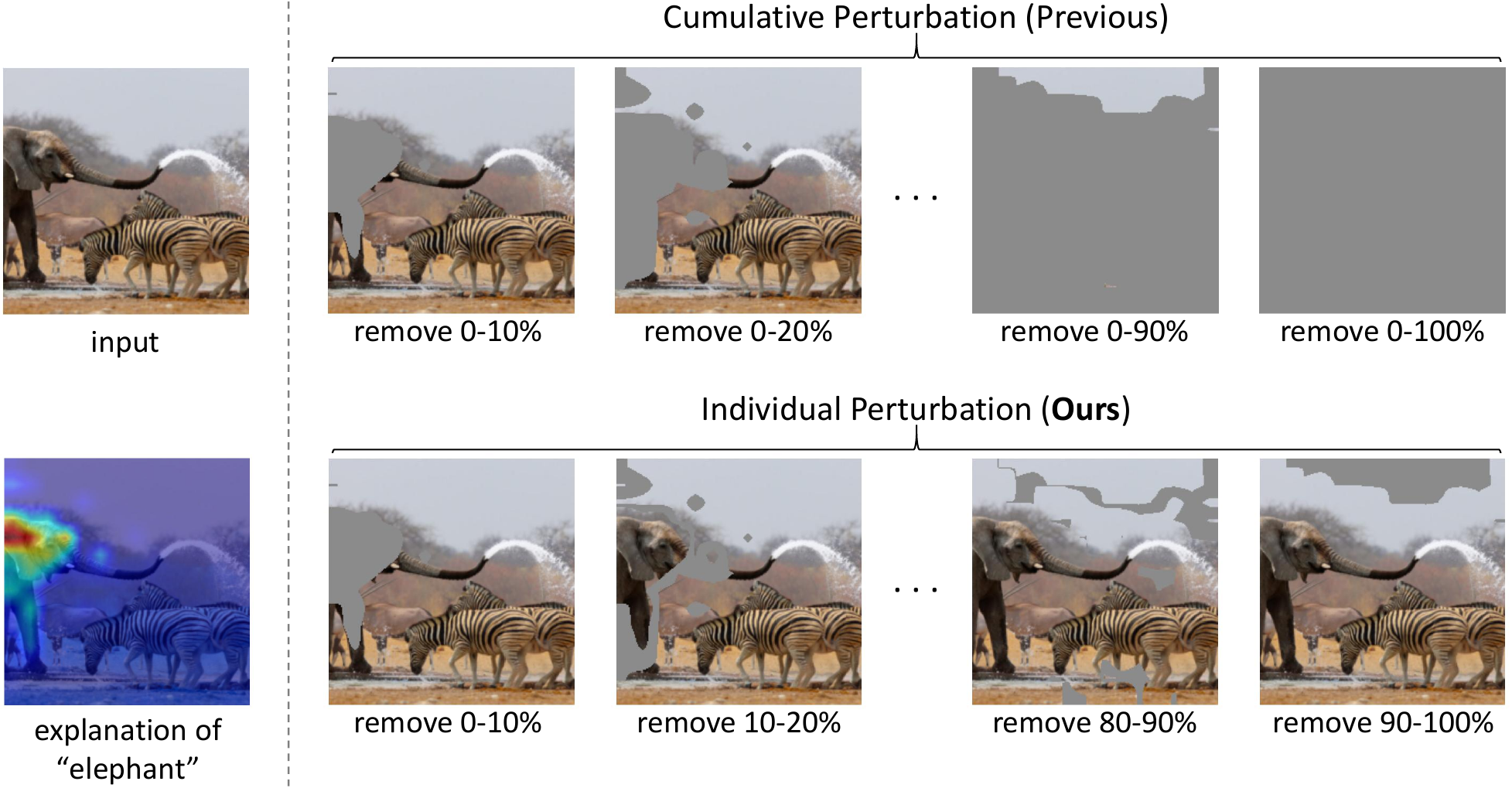}
    }
  \caption{Explanation result and illustration of two perturbation manners: cumulative perturbation and our SaCo perturbation. Previous metrics perturb the pixel subsets cumulatively. In contrast, the SaCo perturbs them individually to directly compare their influences.}
  \vspace{-0.3cm}
  \label{perturbation}
\end{figure*}

In response to these desiderata, for a thorough faithfulness metric, it is necessary to:
\textbf{(i)}
explicitly compare the influences of input pixels with different magnitudes of salience, and
\textbf{(ii)}
quantify the differences in salience scores to reflect the expected disparities in their impacts.
However, existing metrics fall short in both aspects, as they rely on cumulative perturbation \cite{shah2021input} and do not consider the information embedded in the distribution of the magnitude of salience scores.
For example, with cumulative perturbation (see Figure \ref{perturbation}), it is difficult to discern the influence of pixels ranked between the top 0-10\% (the elephant's body) and 90-100\% (the sky) of salience.
This is because the removal of the top 90-100\% important pixels is only performed after the top 0-90\% have been eliminated, which conflates their impacts.
Moreover, without considering the exact values of salience, it is uncertain to what degree an explanation expects pixels in the top 0-10\% to have more influence than those in 90-100\%.
Existing metrics cannot adequately evaluate an explanation's ability to differentiate importance levels among different pixels, thereby failing to validate the faithfulness core assumption.
Such deficiency can lead to unreliable outcomes, underlining the need for a nuanced evaluation. For instance, it is alarming that commonly used metrics fail to distinguish between some state-of-the-art explanation methods and Random Attribution \cite{wang2022unified}.

Recognizing that faithfulness is essential for explanation methods to depict models' behavior, we propose a novel evaluation framework, \textbf{Sa}lience-guided Faithfulness \textbf{Co}efficient, or \textbf{SaCo}, which analyzes how faithfully an explanation method aligns with the model's behavior.
The proposed metric operates by conducting a statistical analysis of pixel subsets with varying salience scores and comparing their impacts on the model's prediction.
The salience score distribution is evaluated based on its alignment with the true effect of corresponding pixels.
For instance, if a pixel subset with higher salience scores significantly impacts the model's prediction more than a subset with lower scores, as anticipated, such a pair of subsets is deemed to satisfy the faithfulness criterion. Consequently, the disparity in salience scores between these two subsets, which represents the degree of expectation, will be positively accumulated to the measured outcome.
Conversely, if a pair of subsets does not meet this expectation, it is identified as a violator and will have a negative contribution to the outcome.
The SaCo is suitable for testing the core assumption validity, as it involves explicit comparisons among different pixels and captures the expected disparities in their impacts.

Experimental results across a range of datasets and Vision Transformer models in Section \ref{exp result} demonstrate that current metrics ignore proper evaluations of faithfulness.
Furthermore, we observe that most explanation methods for Vision Transformers actually underperform when tested against the core assumption.
To investigate the key factors that affect faithfulness, we perform ablative experiments on attention-based explanation methods.

In summary, we state our contributions as follows:
\textbf{(i)}
We develop a new evaluation metric, SaCo, to assess how well explanations adhere to the core assumption of faithfulness. By comprehensive examination of ten representative explanation methods across three datasets and three Vision Transformer models, we demonstrate that SaCo can provide a complementary tool for evaluating how salience scores signify the level of anticipated impacts on the model.
\textbf{(ii)}
Empirically, we demonstrate SaCo's capability to distinguish meaningful explanations from Random Attribution, setting a useful and robust benchmark.
\textbf{(iii)}
We provide insights into certain designs in current attention-base explanation methods that may alter faithfulness. Our empirical results highlight the role of gradient information and aggregation rules, guiding the path for future improvements in Vision Transformer interpretability methodology.

%% file: sec/2_relatedwork.tex
\section{Related Work}
\subsection{Post-hoc Explanations}
\textbf{Traditional post-hoc explanations.}
Traditional post-hoc explanation methods largely fall into two groups: gradient-based and attribution-based approaches. The first group includes methods like Input $\odot$ Gradient \cite{shrikumar2016not}, SmoothGrad \cite{smilkov2017smoothgrad}, Full Grad \cite{srinivas2019full}, Integrated Gradients \cite{sundararajan2017axiomatic}, and Grad-CAM \cite{selvaraju2017grad}. These methods leverage gradient information to calculate salience scores. In contrast, attribution-based methods \cite{bach2015pixel, shrikumar2017learning, nam2020relative, gur2021visualization} propagate classification scores backward to the input and then use the resultant values as indicators of contribution. Beyond these two types, there are also other methods, such as saliency-based \cite{zhou2016learning, mahendran2016visualizing, zhou2018interpreting}, Shapley additive explanation \cite{lundberg2017unified}, and perturbation-based methods \cite{fong2017interpretable, fong2019understanding}. Although initially designed for MLPs and CNNs, some of them have been successfully adapted for Vision Transformers in recent works \cite{chefer2021transformer, ali2022xai}.

\noindent\textbf{Leveraging attentions to interpret Vision Transformers.}
A distinct branch of research in post-hoc interpretability is committed to creating new paradigms specifically for Transformers \cite{Wu_2024_CVPR}. Attention maps are widely used in methodologies associated with this direction, as they inherently constitute distributions that represent the sampling weights over input tokens.
Representative methods include Raw Attention \cite{wiegreffe2019attention} and Rollout \cite{abnar2020quantifying}, which regard attention maps as an explanation, Transformer-MM \cite{chefer2021generic}, a general explanation framework utilizing gradient information, and ATTCAT \cite{qiang2022attcat}, which formulates Attentive Class Activation Tokens to estimate the relative importance among input tokens.
The salience scores produced by these methods are subsequently mapped onto the image space, generating visualizations that are easily understood by humans.
However, whether these interpretation maps are faithful to the model's actual behavior remains a matter of debate \cite{jain2019attention, wiegreffe2019attention, deyoung2020eraser}.

\subsection{Evaluation of Explanation Faithfulness}
In this paper, we evaluate the faithfulness of Vision Transformer explanation methods. This is a critical task \cite{adebayo2018sanity}, especially given the recent debates concerning whether parameters and features in Transformer models are explainable. For instance, the reliability of attention weights has been questioned in several studies \cite{jain2019attention, serrano2019attention, kobayashi2020attention, bastings2020elephant}, premised on the hypothesis that attention weights should not conflict with gradient rankings or token norms. Following this, \cite{wiegreffe2019attention} proposed alternative testing strategies and argued that attention is interpretable when limited to certain circumstances.
However, we find that current studies do not sufficiently consider the values of salience scores and the model's confidence in its original predictions, which deviates from the core assumption of faithfulness and leads to inconsistent and untrustworthy outcomes.

Evaluating post-hoc explanations poses a challenge in the realm of Transformer interpretability. Given the absence of justified ground truth \cite{agarwal2022openxai}, one stream of research focuses on developing human-centered evaluations \cite{lage2018human, ross2018improving, colin2022cannot}. These studies scrutinize the practical value of explanations provided to the end user.
In another direction, \cite{adebayo2018sanity} employed sanity checks to assess changes in explanations \emph{w.r.t.} randomization within models and datasets.
\cite{ancona2017towards} formulated gradient-based explanations in a unified framework and introduced Sensitivity-n as a desired property.
However, this metric highly relies on the linearity assumption for simple DNNs and neglects the order of salience scores.
Unlike these studies, our work is broadly related to faithfulness metrics that evaluate explanations by monitoring the model's performance on perturbed images \cite{atanasova2020diagnostic}.
Various versions of perturbation-based metrics have been introduced, providing measures of input feature impact \cite{shrikumar2017learning, nguyen2018comparing, chen2020generating, deyoung2020eraser}.
Despite the achievements, these approaches use cumulative perturbation without individually contrasting input pixels of varying importance levels.
Furthermore, they do not directly incorporate the information from specific magnitudes of salience scores, focusing only on their relative ordering.
These oversights contribute to deficiencies present in existing metrics.
Our approach, on the other hand, scrutinizes the model's response to each distinct pixel group and sheds light on the relevance of the salience scores, providing a more comprehensive evaluation of explanation faithfulness.

%% file: sec/3_method.tex
\section{Methodology}
For the image classification task, each input is an image comprised of $HW$ pixels. Given an input image $\mathbf{x}$ and the corresponding predicted class $\hat{y}(\mathbf{x}) \in \{1, 2, ..., C\}$, where $C$ is the number of classes under consideration, post-hoc explanations generate a salience map $\mathbf{M}(\mathbf{x}, \hat{y}) \in \mathbb{R}^{HW}$. The value of each entry in $\mathbf{M}(\mathbf{x}, \hat{y})$ ought to reflect the contribution of the corresponding pixel to the model's output.
However, the reliability of these interpretation results remains questionable. This underscores the necessity for further examination of faithfulness.

Following the faithfulness core assumption, the property being investigated is the \emph{extent} to which these salience scores are faithful to the model's actual behavior.
Therefore, our proposed evaluation is designed to assess how effectively the disparity in salience scores signifies the variation in their influences on the model's confidence. Considering a sample $\mathbf{x}$, we reorder the input pixels based on their estimated salience and partition them into $K$ equally sized pixel subsets: $G_1, G_2, ..., G_K$. Each subset $G_i$ comprises pixels with top salience ranking from $(i-1)\frac{HW}{K}$ to $i\frac{HW}{K}$ \cite{nguyen2018comparing, chen2020generating}. Regarding each $G_i$ as a basic unit, we can define the salience of a pixel subset:
\begin{equation}
s(G_i) = \sum_{p \in G_i}\mathbf{M}(\mathbf{x},\hat{y})_p, \quad \text{where} \quad i = 1, 2, ..., K.
\label{subset importance}
\end{equation}
In essence, the salience of a subset $G_i$ is the sum of salience scores over all pixels in $G_i$. Following the convention in literature \cite{shah2021input, chefer2021transformer, wang2022unified}, we adopt a proxy measure to access the model's behavior: we replace pixels that belong to a certain subset with the per-sample mean value \cite{hooker2019benchmark} and then observe the resulting effect on the model's confidence. Formally, we represent the replacement result by $Rp(\mathbf{x}, G_i)$. Therefore, the alterations in the model's prediction can be formulated as follows:
\begin{equation}
\nabla pred(\mathbf{x}, G_i) = p(\hat{y}(\mathbf{x})|\mathbf{x}) - p(\hat{y}(\mathbf{x})|Rp(\mathbf{x}, G_i)),
\label{change in confidence}
\end{equation}
where $Rp(\mathbf{x}, G_i)$ represents the perturbed image, in which pixels in subset $G_i$ are replaced with the per-sample mean value.
The fundamental principle underpinning SaCo is that a subset $G_i$ of higher salience should exert more effects compared to a subset $G_j$ of significantly lower salience. Specifically, if $s(G_i) \geq s(G_j)$, we expect the following inequality to be upheld:
\begin{equation}
\nabla pred(\mathbf{x}, G_i) \geq \nabla pred(\mathbf{x}, G_j).
\label{inequality}
\end{equation}
As the difference between $s(G_i)$ and $s(G_j)$ expands, our expectation for Inequality \eqref{inequality} to hold will intensify. Following this, the growing difference in salience scores should accentuate its influence on the evaluation result. For example, a violation of Inequality \eqref{inequality} should be penalized more when the difference in salience becomes larger, thus better reflecting the deviation from the expected model behavior.

Inspired by the Kendall $\tau$ statistic \cite{kendall1938new}, for a thorough analysis, we look into all possible pairs of $G_i$ and $G_j$ and assess their compliance with faithfulness property.
The assessment is guided by a salience-aware violation test based on inequality \eqref{inequality}.
Concretely, when this inequality is violated, the difference in salience will negatively impact the evaluation result. On the contrary, when the inequality holds true, the salience difference will add positively to the outcome.
For example, suppose for a pair of pixel subsets $G_i$ and $G_j$ with $s(G_i) \geq s(G_j)$, we find $\nabla pred(\mathbf{x}, G_i) < \nabla pred(\mathbf{x}, G_j)$.
Then, the difference in salience, $s(G_i) - s(G_j)$, is considered a penalty that reflects the magnitude of our unfulfilled expectations and will be subtracted from the overall coefficient.
If we observe $\nabla pred(\mathbf{x}, G_i) \geq \nabla pred(\mathbf{x}, G_j)$ as expected, the difference $s(G_i) - s(G_j)$ will serve as a reward and positively contribute to the evaluation outcome.
Detailed steps are elaborated in Algorithm \ref{metric}.

\begin{algorithm}[t]
\caption{Salience-guided Faithfulness Coefficient}
\begin{algorithmic}[1]
\State \textbf{Input:} Pre-trained model $\Phi$, explanation method $\mathcal{E}$, input image $\mathbf{x}$.
\State \textbf{Output:} Faithfulness coefficient $F$.
\State \textbf{Initialization:} $F \leftarrow 0$, $totalWeight \leftarrow 0$
\State Compute the salience map $\mathbf{M}(\mathbf{x}, \hat{y})$ based on $\Phi$, $\mathcal{E}$, and $\mathbf{x}$. Generate $G_i$ and obtain corresponding $s(G_i)$ and $\nabla pred(\mathbf{x}, G_i)$, for $i = 1, 2, ..., K$.
\For{$i = 1$ to $K - 1$}
    \For{$j = i + 1$ to $K$}
        \If{$\nabla pred(\mathbf{x}, G_i) \geq \nabla pred(\mathbf{x}, G_j)$}
            \State $weight \leftarrow s(G_i) - s(G_j)$
        \Else
            \State $weight \leftarrow -(s(G_i) - s(G_j))$
        \EndIf
        \State $F \leftarrow F + weight$
        \State $totalWeight \leftarrow totalWeight + \lvert weight \rvert$
    \EndFor
\EndFor
\State $F \leftarrow F \slash totalWeight$
\State \textbf{Return} $F$
\end{algorithmic}
\label{metric}
\end{algorithm}

As per its definition, the SaCo produces a faithfulness coefficient, denoted as $F$, that ranges from $[-1, 1]$. The sign of $F$ reveals the direction of correlation, \emph{i.e.}, it evaluates if the input pixels with higher salience scores generally exhibit greater or lesser predictive influence on the model. Beyond just the direction, the absolute value of $F$ quantitatively measures the degree of correlation.

%% file: sec/4_experiment_setup.tex
\section{Experimental Setup}\label{setup}
\subsection{Datasets and Models}
We utilize three benchmark image datasets: CIFAR-10, CIFAR-100 \cite{krizhevsky2009learning}, and ImageNet (ILSVRC) 2012 \cite{russakovsky2015imagenet}. Details regarding the scales of data, numbers of classes, and image resolutions for each dataset are provided in the supplementary.
Furthermore, to ensure the reliability of our evaluation, we experiment with three Vision Transformer models that are widely adopted in this field: ViT-B, ViT-L \cite{dosovitskiy2020image}, and DeiT-B \cite{touvron2021training}.
In these models, images are divided into non-overlapping $16\times16$ patches, then flattened and processed to create a token sequence.
For classification, a special token $[CLS]$ is added to the sequence, similar to BERT \cite{devlin2018bert}.

\subsection{Explanation Methods}
We investigate ten representative post-hoc explanation methods spanning three categories, \emph{i.e.}, gradient-based, attribution-based, and attention-based. Each method holds unique assumptions about the network architecture and the information flow.
For a better assessment, selected methods are widely recognized in the explainability literature and also compatible with Vision Transformer models under consideration.
Detailed descriptions of these techniques are provided in the supplementary.

\noindent\textbf{Gradient-based methods.}
We select two state-of-the-art explanation methods from this category: Integrated Gradients \cite{sundararajan2017axiomatic} and Grad-CAM \cite{selvaraju2017grad}. Note that the Grad-CAM method was initially designed for visualizing intermediate features in CNNs. Our implementation follows the prior study in Vision Transformer interpretability \cite{chefer2021transformer}.

\noindent\textbf{Attribution-based methods.}
Unlike gradient-based, attribution-based methods explicitly model the information flow inside the network. We select LRP \cite{binder2016layer}, Partial LRP \cite{voita2019analyzing}, Conservative LRP \cite{ali2022xai}, and Transformer Attribution \cite{chefer2021transformer}) in our experiment for a thorough analysis.

\noindent\textbf{Attention-based methods.}
Regarding the attention-based methods, we employ four variants: Raw Attention \cite{jain2019attention}, Rollout \cite{abnar2020quantifying}, Transformer-MM \cite{chefer2021generic}, and ATTCAT \cite{qiang2022attcat} in our experiments. These methods are specifically designed for the Transformer models.

\subsection{Evaluation Metrics} \label{Metrics}
We compare our proposed SaCo with widely adopted existing metrics to validate its reliability.

\noindent\textbf{Area Under the Curve (AUC) $\downarrow$.}
This metric calculates the Area Under the Curve (AUC) corresponding to the model's performance as different proportions of input pixels are perturbed \cite{atanasova2020diagnostic}. To elaborate, we first generate new data by gradually removing pixels in increments of 10\% (from 0\% to 100\%) based on their estimated salience scores. The model's accuracy is then assessed on these perturbed images, resulting in a sequence of accuracy measurements. The AUC is subsequently computed using this sequence. A lower AUC indicates a better explanation.

\noindent\textbf{Area Over the Perturbation Curve (AOPC) $\uparrow$.}
Rather than measuring the model's accuracy, AOPC \cite{nguyen2018comparing, chen2020generating} quantifies the variations in output probabilities \textit{w.r.t.} the predicted label after perturbations. A higher AOPC indicates a better explanation.

\noindent\textbf{Log-odds score (LOdds) $\downarrow$.}
The LOdds \cite{shrikumar2017learning, qiang2022attcat} evaluates if the pixels considered important are enough to sustain the model's prediction, which is measured on the logarithmic scale. To facilitate fair and reliable comparisons, we gradually eliminate the top 0\%, 10\%, ..., 90\%, and 100\% of pixels, based on their salience scores. This removal process aligns with that employed for calculating AUC and AOPC. A lower LOdds indicates a better explanation.


\noindent\textbf{Comprehensiveness (Comp.) $\downarrow$.}
The Comprehensiveness \cite{deyoung2020eraser} measures if pixels with lower salience are dispensable for the model's prediction. For consistent comparisons, we cumulatively eliminate pixels in the least important 0\%, 10\%, ..., 90\%, and 100\%. A lower Comprehensiveness indicates a better explanation.

%% file: sec/5_experiment_result.tex
\section{Experimental Results}\label{exp result}
\subsection{Interrelationships among Evaluation Metrics}

\begin{figure}[t]
  \centering
    \resizebox{0.75\columnwidth}{!}{%
  \includegraphics[width=\columnwidth]{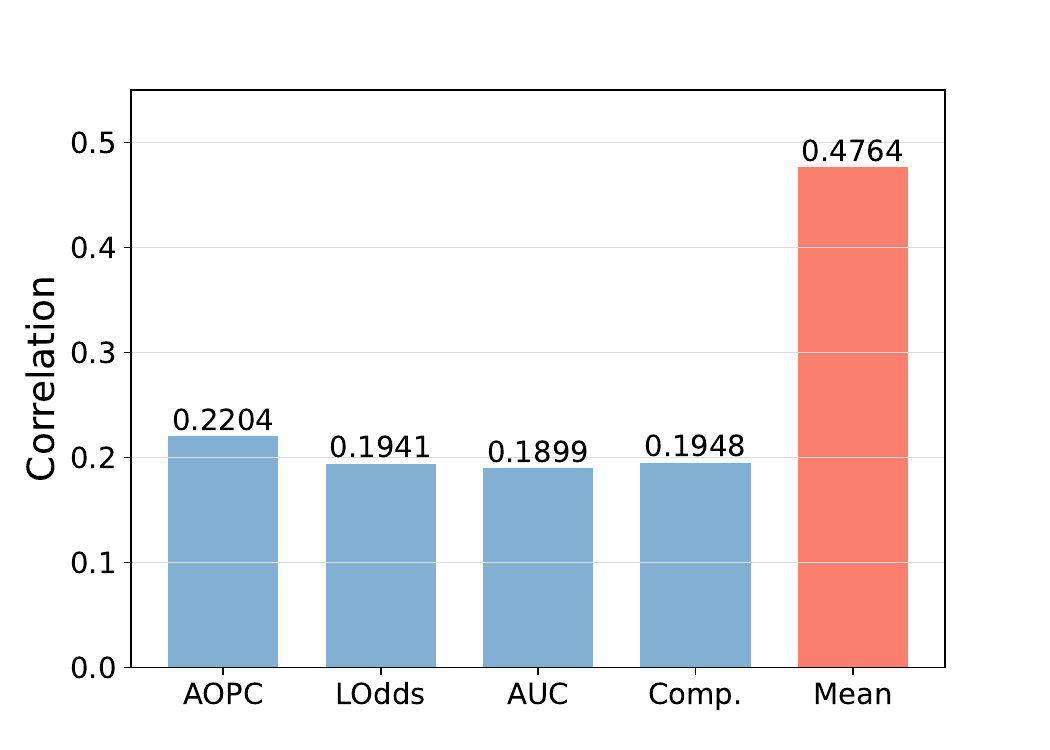}
  }
  \caption{Correlations between sample rankings \emph{w.r.t.} our SaCo and existing metrics.}
  \label{correlation}
  \vspace{-0.4cm}
\end{figure}

To demonstrate the significance and necessity of SaCo, we conduct a correlation analysis following the thorough experimental setup.
To this end, we begin by evaluating each explanation method on single samples independently, using each of the metrics.
Then, we compute the statistical rank correlations between the evaluation results obtained from the SaCo and those from other existing metrics.
Note that we are correlating based on the rankings rather than the exact values of evaluation results, as these metrics can vary in scales and orientations.
This approach allows us to quantify the degree of similarity between the assessments provided by the SaCo and the traditional metrics in current use.
Figure \ref{correlation} shows the comprehensive statistical correlation results averaged across all considered datasets, explanation methods, and Vision Transformer models, as described in Section \ref{setup}.
The first four bars on the left depict the correlations between our SaCo and the metric indicated by the corresponding x-axis label. The rightmost bar shows the average correlation among all other metrics, excluding ours.

As displayed, the correlation scores between our SaCo and other existing metrics range from 0.18 to 0.22 (in this analysis, a result of 1 indicates a complete correlation, while a result of 0 indicates no correlation).
These low scores signify minimal congruence in their evaluation,
suggesting that our SaCo potentially evaluates a complementary aspect compared to existing metrics, \emph{i.e.}, the core assumption of faithfulness.
In essence, the traditional metrics tend to generate similar results regardless of the degree of faithfulness, due to their lack of comparisons among individual pixels and the direct incorporation of the distribution of salience score magnitudes.
On the other hand, the average intra-correlation among the existing metrics themselves is significantly higher (0.4764). This result implies that these metrics tend to evaluate similar or overlapping aspects of interpretations (mainly the effect of progressive pixel removal), with insufficient consideration of the faithfulness assumption.
These results emphasize the importance of our proposed SaCo, as current metrics appear to lack the capabilities to adequately assess faithfulness.
Therefore, the need for a more comprehensive assessment method for post-hoc explanations of Vision Transformers is reinforced.

\begin{figure*}[t]
\centering
    \resizebox{0.83\textwidth}{!}{%
  \includegraphics[width=0.83\textwidth]{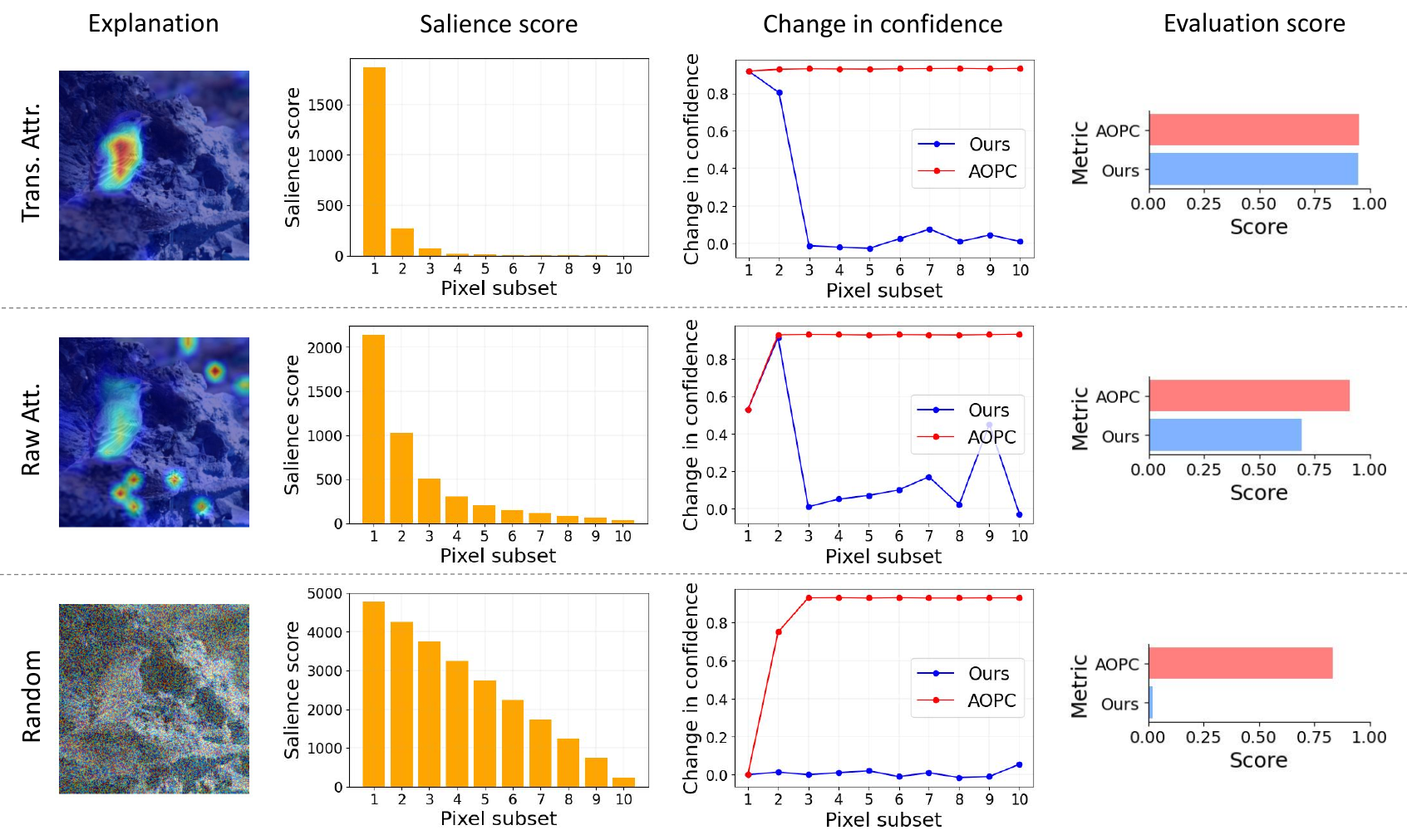}
  }
\vspace{1mm}
\caption{Illustration of three explanations for the predicted class `linnet', salience score distributions, changes in model's confidence caused by perturbation, and final SaCo and AOPC scores.}
  \vspace{-0.3cm}
  \label{casestudy}
\end{figure*}

\subsection{Evaluating Random Attribution}
Recognizing the potential shortages of current metrics in assessing faithfulness, we conduct a critical examination to further demonstrate the limitations: we evaluate the performance of Random Attribution as an explanation method \cite{hooker2019benchmark, shah2021input}.
In this context, Random Attribution directly assigns salience scores to input pixels using a purely uniform distribution,
with no reference to the information of the model's internal inference process \cite{wang2022unified}.
From the perspective of mathematical expectation, such Random Attribution represents a complete absence of faithfulness \cite{shah2021input}, as the assigned salience scores bear no discriminative relationship to the actual impacts of the pixels on the model.
As a result, an ideal metric of faithfulness is expected to return a distinct score on Random Attribution, essentially setting a baseline indicative of a significant lack of meaningful understanding of the model's prediction.
In particular, from the viewpoint of mathematical expectation, our SaCo yields a faithfulness coefficient of zero for Random Attribution.
Given that the salience scores are sampled from a uniform distribution, the pixels subsets ($G_i, i = 1, 2, ..., K$) partitioned according to these scores will not show significant differences in their discriminative information.
Therefore, Inequality \eqref{inequality} holds true with a probability of one-half, leading to a balance between the positive and negative terms during the calculation of the coefficient outlined in Algorithm \ref{metric}.

\noindent\textbf{Case study.}
We first present a case study comparing the behavior of SaCo and AOPC.
AOPC operates on cumulative pixel perturbation and disregards the alignment between salience values and actual influences.
\begin{figure*}[t]
\vspace*{-3mm}
\centering
\begin{minipage}{0.329\textwidth}
  \centering
    \resizebox{\textwidth}{!}{%
  \includegraphics[width=\textwidth]{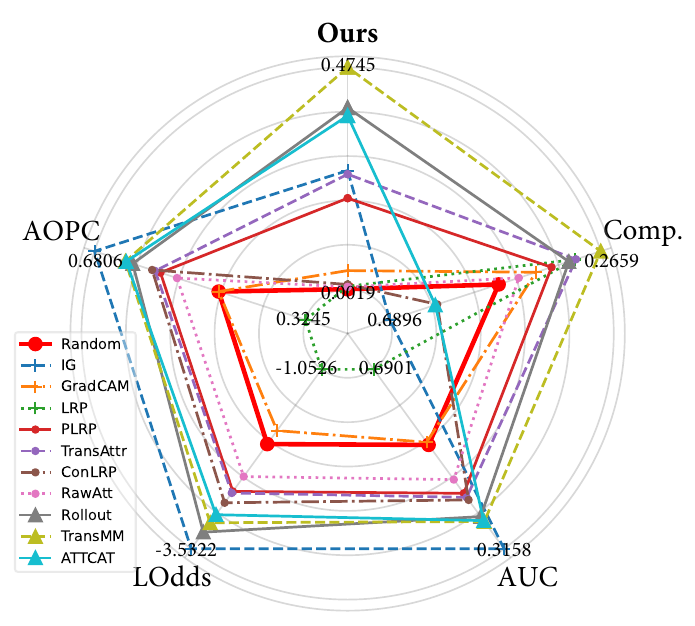}
  }
\end{minipage}
\begin{minipage}{0.329\textwidth}
  \centering
    \resizebox{\textwidth}{!}{%
  \includegraphics[width=\textwidth]{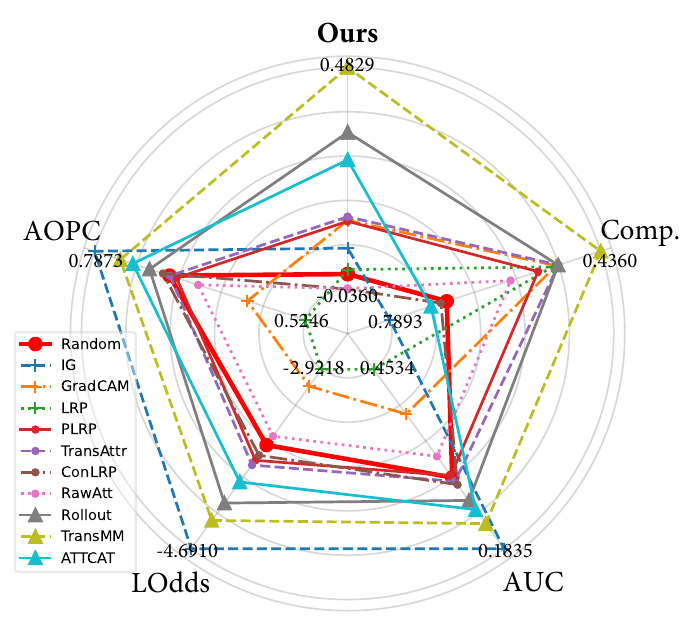}
  }
\end{minipage}
\begin{minipage}{0.329\textwidth}
  \centering
    \resizebox{\textwidth}{!}{%
  \includegraphics[width=\textwidth]{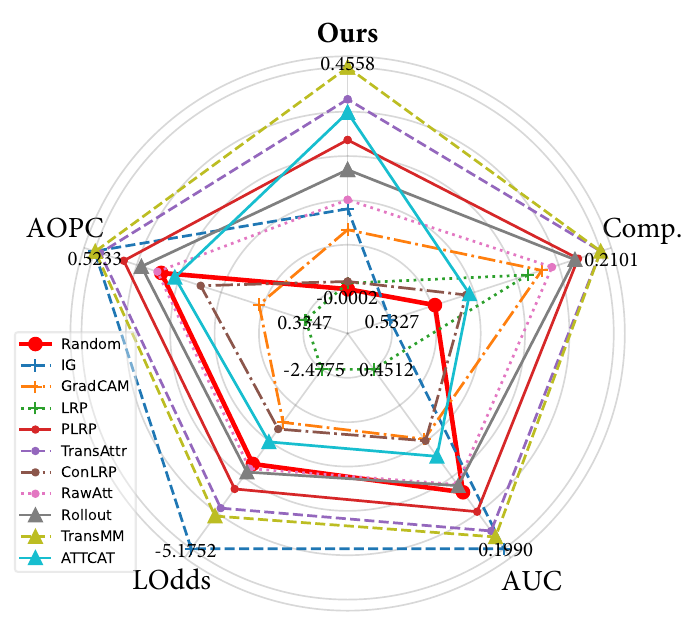}
  }
\end{minipage}
\vspace{1mm}
\caption{Evaluation results for advanced explanation methods and Random Attribution (\textcolor{red}{red}). Three graphs present results on CIFAR-10 (left), CIFAR-100 \cite{krizhevsky2009learning} (middle), and ImageNet \cite{russakovsky2015imagenet} (right), respectively. The values on each axis have been rescaled so that a larger distance from the center consistently signifies superior performance. Enlarged graphs are provided in the supplementary for better clarity.}
\vspace{-0.3cm}
\label{random}
\end{figure*}
We conduct this study on ViT-B \cite{dosovitskiy2020image} and a sample from ImageNet \cite{russakovsky2015imagenet}.
As shown in Figure \ref{casestudy}, we evaluate three explanation methods: Transformer Attribution \cite{chefer2021transformer}, Raw Attention, and Random Attribution. Specifically, following the literature convention \cite{selvaraju2017grad, chefer2021transformer, chefer2021generic, pan2021ia, zheng2022shap}, we divide the image into ten disjoint pixel subsets, designated by $K{=}10$. The salience score of each subset $s(G_i)$ is computed by Eq. \eqref{subset importance}. We then implement both SaCo's individual perturbation and AOPC's cumulative perturbation on the images, and calculate the resulting changes in the model's confidence (probability) for the predicted class, as defined by Eq. \eqref{change in confidence}.
Here, a positive change represents a decrease in confidence.

Across all methods, the confidence drops induced by cumulative perturbation remain almost unchanged (above 0.9) after the removal of the top 30\% important pixels, despite the subsequent removal of more pixels.
This pattern results in consistently high AOPC scores for all three methods.
Furthermore, one can observe that the impacts of cumulative pixel removal do not increase linearly, which holds true even for uniform Random Attribution.
The precipitous decline in confidence may stem from the out-of-distribution issues that arise due to the substantial removal of pixels \cite{hooker2019benchmark, shah2021input, hase2021out}.
As discussed in Section \ref{intro}, in cumulative perturbation, less important pixels are removed only after the most important pixels have been eliminated, causing their individual effects to become entwined.
Conversely, SaCo directly assesses the influences of individual pixel subsets and measures their alignment with the salience distribution.
In the case of Transformer Attribution, the influences of pixel subsets (indicated by the blue curve) closely align with the salience score distribution, yielding a high SaCo result. In contrast, with Raw Attention, the subsets $G_i (i = 3, 4, 5, 6)$ have minimal impacts on the model, possibly due to unexpected emphasis on irrelevant objects and backgrounds, as reflected by a reduced SaCo score. For Random Attribution, the influences of subsets exhibit little variation, resulting in a near-zero SaCo score.
This case study shows that SaCo provides a rigorous evaluation that effectively differentiates superior and inferior explanations.

\noindent\textbf{Large-scale experiments.}
To further validate the effectiveness of SaCo, we conduct experiments on large-scale datasets.
Figure \ref{random} presents the overall outcomes on CIFAR-10, CIFAR-100 \cite{krizhevsky2009learning}, and ImageNet \cite{russakovsky2015imagenet}.
For every dataset, we compare existing metrics with SaCo, where each result is an average over three Vision Transformers indicated in Section \ref{setup}.
Across all datasets, SaCo consistently scores Random Attribution near zero, while most explanation methods obtain positive scores under SaCo, demonstrating its ability to set a standard benchmark.
Conversely, other metrics cannot provide consistent evaluation for Random Attribution.
When assessed using these metrics, despite producing purely noisy heatmaps, Random Attribution appears to have comparable performance, even surpassing some state-of-the-art methods such as Partial LRP \cite{voita2019analyzing}, Transformer Attribution \cite{chefer2021transformer}, and ATTCAT \cite{qiang2022attcat}, as demonstrated in Figure \ref{random}.
Another observation is that existing metrics seem to be sensitive to the removal order of the cumulative perturbation they employ.
A phenomenon across three datasets in our experiment exemplifies this problem: while I.G. performs best on AOPC, LOdds, and AUC (with Most Relevant First removal), it is surprisingly the worst on Comprehensiveness (with the reverse removal order).
This indicates that current metrics are significantly inconsistent \emph{w.r.t.} hyperparameters such as removal orders \cite{rong2022consistent}.
In contrast, we eliminate this inconsistency by directly comparing individual pixel subsets of various salience scores, instead of cumulatively removing them in a specific order.

\subsection{Assessment of Current Explanation Methods}
Figure \ref{random} illustrates the results of SaCo for existing explanation methods over Vision Transformer models and datasets under consideration.
Observations from our results indicate that all explanation methods perform moderately, as reflected in their suboptimal SaCo scores.
These results necessitate in-depth research into explanation methods to more accurately depict the model's reasoning process and adhere to the core assumption of faithfulness.
Furthermore, of all explanation methods evaluated, we can see that those utilizing attention information generally perform better in our SaCo assessment.
However, despite falling under the same category, Raw Attention noticeably underperforms. This motivates us to hypothesize that attention-based explanation methods can only achieve superior performance when they incorporate auxiliary information, such as gradient and cross-layer integration.
We further conduct experiments in Section \ref{designs} for its demonstration.

\begin{table}[b]
\centering
\vspace{-0.5cm}
\resizebox{0.65\columnwidth}{!}{%
\begin{tabular}{c|c|c}
\toprule
cross-layer aggregation & gradient & SaCo $\uparrow$ \\
\hline
\xmark & \xmark & 0.1835\\
\checkmark & \xmark & 0.2453\\
\xmark & \checkmark & 0.3783\\
\checkmark & \checkmark & \textbf{0.4558}\\
\bottomrule
\end{tabular}
}
  \vspace{0.1cm}
\caption{Ablative study on attention-based explanation methods.}
\label{ablation}
  \vspace{-0.3cm}
\end{table}

\begin{table*}[t]
\centering
    \resizebox{1\linewidth}{!}{%
\begin{tabular}{c|cccccccccc}
\toprule
K&I. G. \cite{sundararajan2017axiomatic}&Grad-CAM \cite{selvaraju2017grad}&LRP \cite{binder2016layer}&P. LRP \cite{voita2019analyzing}&Trans. Attr. \cite{chefer2021transformer}&Con. LRP \cite{ali2022xai}&Raw Att. \cite{jain2019attention}&Rollout \cite{abnar2020quantifying}&Trans. MM \cite{chefer2021generic}&ATTCAT \cite{qiang2022attcat}\\
\hline
5&0.1585&0.1659&0.0246&0.4628&0.5680&0.0544&0.3200&0.3372&0.6041&0.3178\\
10&0.1647&0.1142&0.0120&0.3066&0.3902&0.0155&0.1835&0.2453&0.4558&0.3629\\
20&0.1785&0.1000&0.0054&0.2282&0.2906&-0.0201&0.1411&0.1956&0.3651&0.3617\\
\bottomrule
\end{tabular}
}
\vspace{1mm}
\caption{Performance of current explanation methods on our SaCo, with different values of $K$. Results are averaged over three Vision Transformer models on ImageNet \cite{russakovsky2015imagenet}.}
\label{more ablation}
  \vspace{-0.6cm}
\end{table*}

\subsection{Effects of Designs in Explanation Methods}\label{designs}
We now delve into the designs of explanation methods that may augment the alignment with the faithfulness core assumption. We focus our analysis on attention-based explanation methods because attention weights are inherently meaningful for Vision Transformers \cite{chefer2021generic}. Moreover, previous assessments have demonstrated that attention-based methods generally outperform others \cite{chefer2021transformer, qiang2022attcat}, making them a valuable factor for further investigation.

As depicted in Figure \ref{random}, attention-based explanation methods that use well-crafted aggregation rules and auxiliary information from gradient \emph{w.r.t.} the model's output tend to score higher under SaCo.
Based on this observation, we hypothesize that the incorporations of aggregation rules and gradient information play a vital role in compliance with faithfulness.
To validate this hypothesis, we conduct ablative experiments employing four variants of attention-based explanation methods:
\textbf{(i)}
utilizing attention weights in the final layer of the model,
\textbf{(ii)}
aggregating attention information across all layers, 
\textbf{(iii)}
utilizing the final layer's attention weights and integrating their gradient information, and
\textbf{(iv)}
aggregating attention weights across all layers and integrating gradient information.


Table \ref{ablation} presents our ablation study's results on attention-based explanation methods, showing the benefits of integrating gradient information and multi-layer aggregation.
Firstly, we can observe that the incorporation of gradient information significantly improves faithfulness scores. Specifically, when considering only the last layer, the introduction of gradients boosts the evaluation outcome by approximately 106\%.
This effect remains robust, showing an 86\% increase even with the application of aggregation across all layers.
Secondly, despite being less influential than gradient information, aggregating across multiple layers also contributes positively to the results.
This improvement occurs irrespective of whether the gradient is used, suggesting that a more holistic view of the Vision Transformer can consistently facilitate more faithful explanations.
The results empirically support our initial hypothesis that both the gradient and aggregation rules are essential for Vision Transformer explanations, with the former having a more significant effect.
Refining these two design factors offers a promising avenue for advancing the development of explanations for Vision Transformers.
Furthermore, this study also demonstrates SaCo's capability to capture the property of faithfulness.
The result resonates with the intuitive human understanding that a precise depiction of the model's reasoning regarding the recognized object requires class-specific information and comprehensive insights from the entire inference process.

\subsection{Exploring Influential Factors in SaCo}
\noindent\textbf{The number of pixel subsets (K).}
To explore the impact of varying the number of pixel subsets, we assess the performance of existing explanation methods across different values of $K$ in Algorithm \ref{metric}.
Table \ref{more ablation} presents the averaged results across three Vision Transformer models, evaluated by our SaCo with different $K$ values.
It can be observed that with an increase in $K$, most methods exhibit a slight decline in their SaCo scores.
This trend is more pronounced for Partial LRP \cite{voita2019analyzing}, Transformer Attribution \cite{chefer2021transformer}, and Transformer-MM \cite{chefer2021generic}, suggesting that these explanation methods might struggle to maintain faithfulness under a more granular evaluation.
Conversely, Integrated Gradients (I. G.) and ATTCAT show relatively consistent performance across different $K$ values, indicating stronger robustness to the granularity of subset division.
These results emphasize the significance of choosing an appropriate $K$.
The optimal value of $K$ depends on the specific needs of evaluation, striking a balance between computational demands and the granularity of the faithfulness evaluation.

\noindent\textbf{Measure of distinct salience scores.}
In our proposed SaCo (see Algorithm \ref{metric}), we quantify the extent of satisfying or violating human expectation by the differences in salience scores, expressed as $weight \leftarrow s(G_i) - s(G_j)$.
One possible alternative for this measure is taking the ratio: $weight \leftarrow \frac{s(G_i)}{s(G_j)}$, which seems promising because ratios are effective at capturing the relative magnitude of salience among pixel subsets.
However, using a ratio violates the property of scale-invariance.
In practice, the explanation results may be normalized or scaled into $[0, 1]$ interval as post-processing \cite{selvaraju2017grad, chefer2021transformer, chefer2021generic, qiang2022attcat}, which will skew the ratio measure.
This may cause extremely high or even infinite ratios when $s(G_j)$ is close to zero after transformation, which is especially problematic when comparing explanations from different methods that scale their salience scores differently.
In contrast, our designed SaCo maintains the scale-invariance property. Regardless of the scale on which the salience scores are expressed, the results remain consistent. This property enables a stable comparison between distinct methods thereby ensuring a more robust evaluation.
Formal proof demonstrating that our SaCo satisfies the scale-invariance property is provided in the supplementary.

%% file: sec/6_conclusion.tex
\section{Conclusion}
In this work, we proposed SaCo, a novel faithfulness evaluation.
Our SaCo leverages salience-guided comparisons of pixel subsets for their different contributions to the model's prediction, providing a more robust benchmark.
Experiments reveal insightful observations:
\textbf{(i)}
our correlation analysis shows the necessity of SaCo, as existing metrics capture overlapping aspects while lacking consideration of faithfulness,
\textbf{(ii)}
unlike existing metrics, SaCo can identify Random Attribution as completely lacking significant information and provide consistent results free of removal order dependency, and
\textbf{(iii)}
attention-based methods are generally more faithful, and their performance can be further enhanced by gradient information and multi-layer aggregation.
In summary, our work provides a comprehensive evaluation of faithfulness in Vision Transformer explanations, which will spur further research in explainability.

\noindent \textbf{Acknowledgments:} This research is supported by NSF IIS-2309073 and ECCS-212352101. This article solely reflects the opinions and conclusions of its authors and not the funding agencies.